\documentclass[runningheads]{llncs}
\usepackage[T1]{fontenc}
\usepackage{graphicx}
\usepackage{booktabs}
\usepackage[misc]{ifsym}
\newcommand{\corr}{(\Letter)}

\usepackage{cite} 
\usepackage{float} 
\usepackage{acro} 

\DeclareAcronym{OPRO}{
	short = OPRO,
	long = Optimization by Prompting,
}
\DeclareAcronym{LLM}{
	short = LLM,
	long = Large Language Model,
}
\DeclareAcronym{NLP}{
	short = NLP,
	long = Natural Language Processing,
}
\DeclareAcronym{NLI}{
	short = NLI,
	long = Natural Language Inference,
}
\DeclareAcronym{CoT}{
	short = CoT,
	long = Chain of Thought,
}
\DeclareAcronym{CQA}{
	short = CQA,
	long = Commonsense Question-Answering,
}

\begin{document}

\title{Leveraging Zero-Shot Prompting for Efficient Language Model Distillation}

\titlerunning{Leveraging Zero-Shot Prompting for Efficient Language Model Distillation}

\author{
	Lukas Vöge \inst{1} \and
	Vincent Gurgul \inst{1} \corr \and
	Stefan Lessmann \inst{1, 2}
}

\authorrunning{L. Vöge et al.}

\institute{
	Humboldt-Universität zu Berlin, Unter den Linden 6, 10117 Berlin \and
	Bucharest University of Economic Studies, Piața Romană 8, 010374 Bucharest}

\maketitle 


\begin{abstract}

This paper introduces a novel approach for efficiently distilling \acp{LLM} into smaller, application-specific models, significantly reducing operational costs and manual labor. Addressing the challenge of deploying computationally intensive \acp{LLM} in specific applications or edge devices, this technique utilizes \acp{LLM}' reasoning capabilities to generate labels and natural language rationales for unlabeled data. Our approach enhances both finetuning and distillation by employing a multi-task training framework where student models mimic these rationales alongside teacher predictions. Key contributions include the employment of zero-shot prompting to elicit teacher model rationales, reducing the necessity for handcrafted few-shot examples and lowering the overall token count required, which directly translates to cost savings given the pay-per-token billing model of major tech companies' \ac{LLM} APIs. Additionally, the paper investigates the impact of explanation properties on distillation efficiency, demonstrating that minimal performance loss occurs even when rationale augmentation is not applied across the entire dataset, facilitating further reductions of tokens. This research marks a step toward the efficient training of task-specific models with minimal human intervention, offering substantial cost-savings while maintaining, or even enhancing, performance.


\end{abstract}


\section{Introduction}

The rapid advancement of \acp{LLM} such as ChatGPT has facilitated significant progress in \ac{NLP}, offering unprecedented performance across a wide array of tasks. \acp{LLM}, by their nature, embody extensive knowledge bases and reasoning abilities but come with high computational and memory costs, making their direct application in specific real-world tasks inefficient. This challenge has catalyzed the exploration of techniques to distill the essence of \acp{LLM} into smaller, more manageable models through two primary methodologies: finetuning and distillation. Finetuning adjusts pre-trained models to specific tasks using human-labeled data, while distillation transfers knowledge from a large teacher model to a smaller student model, often leveraging the teacher's outputs as training data. Both strategies conventionally require substantial volumes of high-quality data, with finetuning being limited by the availability and cost of human annotations, and distillation grappling with the inherent noise in \ac{LLM}-generated labels.

Addressing these limitations, our work builds upon the innovative approach proposed by \cite{hsieh_2023_distilling}, which utilizes LLMs to generate not only labels for unlabeled data but also natural language rationales that justify these labels. This method, termed step-by-step distillation by the authors, formulates finetuning and distillation as a multi-task process where the student model learns to replicate both the label and the rationale provided by the teacher, significantly enhancing the efficiency and performance of smaller models with less data.

This paper further advances step-by-step distillation by applying zero-shot prompting techniques to procure label-rationale pairs from teacher \acp{LLM}, eliminating the labor-intensive creation of few-shot examples as previously required by \cite{hsieh_2023_distilling}. This approach, leveraging instruction-tuned models like GPT-3.5-turbo, presents a substantial reduction in operational costs, particularly for applications relying on pay-per-token APIs from major tech companies. This cost-efficiency improvement forms the cornerstone of our primary research goal: to enhance the economic viability of step-by-step distillation by minimizing the data queried from teacher \acp{LLM}.

Furthermore, we delve into the effects of rationale properties on the distillation process---a relatively underexplored area in current literature. Our investigations cover the optimal number of dataset instances requiring rationales, the impact of rationale accuracy, and how the length of explanations influences the performance of student models of varying sizes.

With these objectives, our paper contributes to the understanding and application of \ac{LLM} distillation, providing insights into how to leverage \ac{LLM} reasoning capabilities more efficiently. Through a structured exploration covering both the cost-efficient employment of zero-shot prompts and the strategic use of rationales, this study unveils practical strategies for enhancing the performance of task-specific models with minimal resource expenditure. Our work starts with an overview of related literature, followed by a presentation of our methodology and experimental design, and ends in a discussion of our findings and their implications for the field.


\section{Related Work}

Step-by-step distillation is a technique that leverages teacher LLMs to generate explanatory rationales alongside predictions, thereby enriching the data used for finetuning smaller models. Central to this methodology, pioneered by \cite{hsieh_2023_distilling}, is the application of \ac{CoT} prompting, as proposed by \cite{wei_2022_chain}. This technique induces \acp{LLM} to generate intermediate reasoning steps in their answer, which is shown to greatly improve the quality of responses to questions that require multi-step logic. In this context there are two approaches: 1) zero-shot prompting, which involves guiding language models with a task description alone, requiring no previous examples, and 2) few-shot prompting, which involves providing a small number of example outputs to teach the model how to approach a task. Few-shot learning has shown superior performance on many tasks compared to zero-shot methods, though some studies present contrary findings, highlighting situations where zero-shot prompting is equally effective \cite{li_2023_a}.

\begin{figure}[H]
	\centering
    \includegraphics[width=\linewidth]{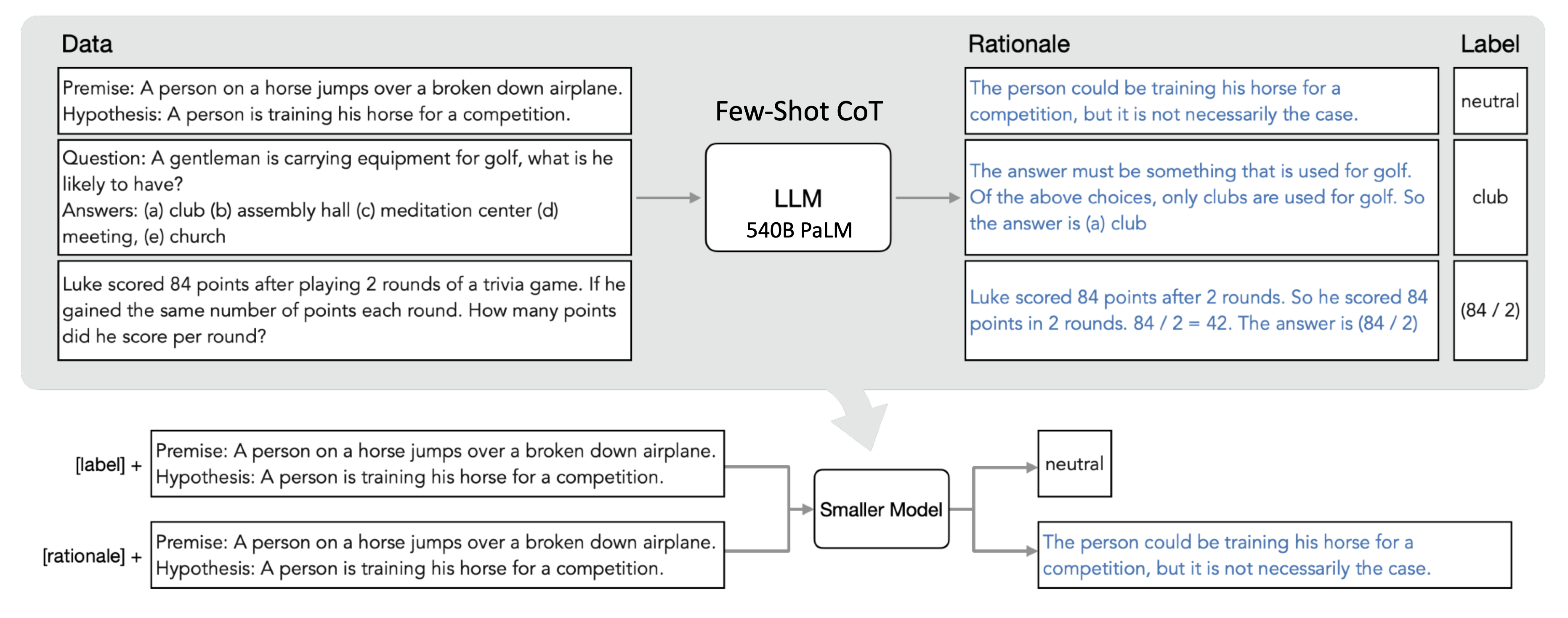}
    \caption{Step-by-step distillation as proposed by \cite{hsieh_2023_distilling}}
	\label{fig:step}
\end{figure}

By employing few-shot \ac{CoT} prompting and a 540B PaLM model as a teacher \cite{hsieh_2023_distilling} achieve enhanced student model performance with fewer data points compared to standard finetuning and distillation on various \ac{NLI} tasks. They finetune T5 models of various sizes on a dual-task framework, optimizing both label accuracy and rationale coherence through a composite of cross-entropy losses. The process of step-by-step distillation is depicted in Fig. \ref{fig:step}.

A similar strategy is adopted by \cite{li_2022_explanations}, focusing on enhancing T5 models for question answering with nuanced \ac{CoT} prompting. However, their methodology necessitates access to ground truth labels for optimizing rationale quality, diverging from \cite{hsieh_2023_distilling} where direct access to labels is not a prerequisite.

\cite{ho_2022_large} introduce a shift towards zero-shot \ac{CoT} prompting as proposed by \cite{kojima_2022_large}, eliminating the reliance on pre-crafted examples and instead exploring a broader interaction with the teacher model to collect a large sample of rationales per dataset instance. This approach, however, leads to increased token consumption due to extensive querying and once again requires access to ground truth labels for rationale pruning. Furthermore, they apply next token prediction, finetuning on complete prompt-rationale-label sequences, which differs from the segmented approach of separately predicting labels and rationales.

\cite{fu_2023_specializing} explores the distillation of multi-step reasoning skills from \acp{LLM} into sub-10B parameter models, focusing on domain-specific capabilities. Utilizing tasks from the GSM8K \cite{cobbe_2021_training} and BigBench Hard \cite{srivastava_2022_beyond} datasets, they investigate the balance between reasoning abilities and model size. Their distillation process, capturing labels from a GPT-3 teacher model in a zero-shot manner, notably does not involve rationale extraction. Their objective centers on minimizing the Kullback-Leibler Divergence between teacher and student model outputs, in turn creating the challenge of accessing the teacher's distribution over its vocabulary. This study validates the potential for smaller models to adopt \ac{CoT} processes.

Unlike the above studies, \cite{li_2023_symbolic} examine the effect of rationale properties on distillation efficiency. They compile a substantial rationale pool for each data point, selectively refining it by excluding inaccuracies, assuming ground-truth label availability. Their investigation into how rationale quantity and diversity influence student model performance underscores the importance of rationale construction for distillation success.

All rationale-based approaches either utilize big few-shot prompts to obtain the required label-rationale pairs or sample between many rationales for each data point. Both of these approaches imply large numbers of tokens that need to be sent and received from the teacher \ac{LLM} and a lot of human labor. Building upon \cite{hsieh_2023_distilling}, our study introduces a novel step-by-step distillation approach based around zero-shot prompting, which facilitates substantial cost-savings by lowering the token requirement and removes the need for any human input. Furthermore, our method does not presuppose access to ground truth labels for rationale validation. We extend the exploration into the impacts of rationale properties on distillation efficiency, by analyzing how rationale length, accuracy, and style affect distilled model performance. This provides several new insights into the efficient application of \ac{CoT} prompting in \ac{LLM} distillation.


\section{Methodology}

Our methodology advances the field of \ac{LLM} distillation by leveraging zero-shot \ac{CoT} prompting to efficiently distill \acp{LLM} into smaller, specialized models. This approach focuses on overcoming the inherent challenges associated with traditional few-shot \ac{CoT} prompting, notably the extensive token usage and the need for elaborate prompt designs that can accurately convey task requirements, style, and formatting in a concise manner.

Our approach is based on instruction-tuned language models, which can generate detailed rationales and label responses even without explicit few-shot examples. While few-shot prompting has shown to effectively guide models to produce responses in a desired format, it often results in lengthy prompts that are counterproductive to our goal of reducing computational costs. Crafting a suitable zero-shot prompt which uses significantly fewer tokens while still eliciting high-quality rationale-label pairs presents a challenge. For example, we must ensure that the model's responses adhere to a predictable format for easy downstream processing, a task complicated by the varied nature of model outputs, even to identical prompts.

To address these challenges, we develop a strategy that not only seeks to minimize the length and increase the efficiency of prompts but also ensures the quality of generated rationales and labels. Our method involves the optimization of zero-shot prompting to identify a template that maximizes the accuracy of generated labels and the inclusion rate of rationales, a critical factor given the sensitivity of models to slight variations in prompt wording and structure. Through this adaptation of step-by-step distillation, we address the main challenges of reducing token usage, minimizing human input and ensuring the generation of predictable, format-consistent model responses.

\begin{figure}[H]
    \centering
    \includegraphics[width=\linewidth]{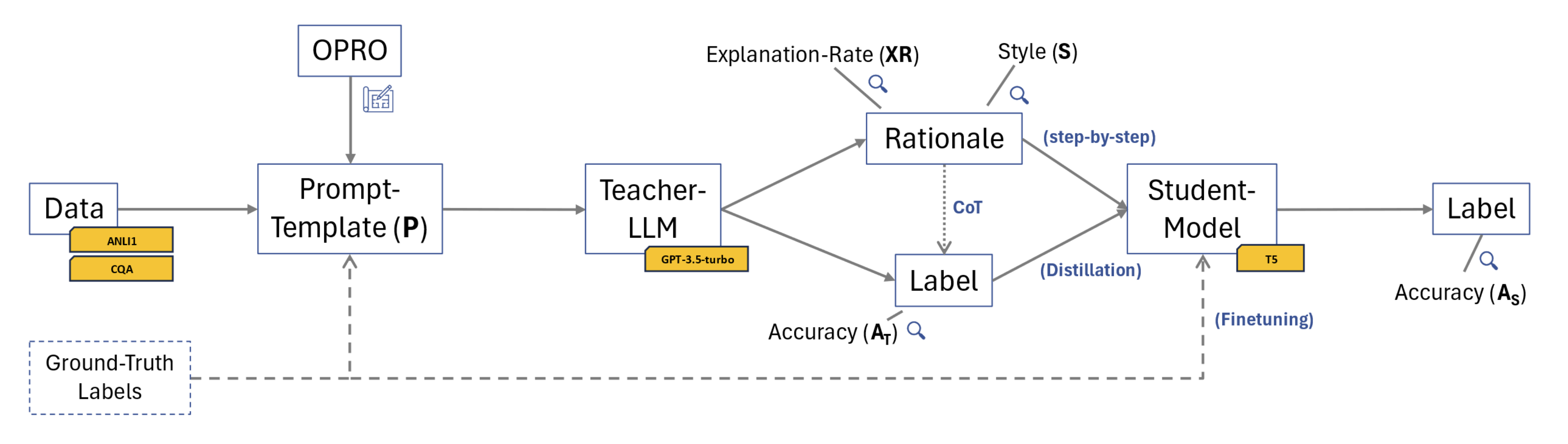}
    \caption{Overview of the proposed zero-shot step-by-step distillation}
\end{figure}

We try to find a zero-shot prompt template $P$ that maximizes the student model accuracy $A_S(A_T(P), \mathit{XR}(P), S(P))$ via step-by-step distillation. Here, $\mathit{XR}$ is the explanation rate, that is, the fraction of teacher responses that include a rationale, and $S$ is the style of the explanation, which includes formatting and length. Since this connection is extremely expensive to measure directly, we approximate the student model accuracy $A_S$ with the teacher \ac{LLM} accuracy $A_T(P)$ and explanation rate $\mathit{XR}(P)$. This is feasible, as \cite{hsieh_2023_distilling} shows, due to the positive correlation between the accuracy of teacher label-rationale pairs and the student model accuracy.

Maximization for $\mathit{XR}(P)$ and $S(P)$ happens implicitly, as \cite{wei_2022_chain} demonstrate in their \ac{CoT} prompting research, since rationales improve the accuracy of \ac{LLM}-predicted labels on tasks that require reasoning. The best performing prompt will therefore also have a high explanation rate. This assumption is additionally supported by our choice of teacher model, which has a very high explanation rate by default, and is confirmed in our experiments. No additional consideration is required to obtain high accuracy prompts with equally high explanation rates. Therefore, our optimization problem simplifies to:

\[ \arg\max_P A_T(P) \]

The process of finding the optimal prompt template $P$ is constrained by two main challenges---the set of all possible prompt templates is infinite, and prompts are discrete. Additionally, the prompt must generate responses that can be parsed into distinct label and rationale parts and minimal human intervention should be required.

We address these challenges by applying the \ac{OPRO} framework \cite{yang_2023_large}, which leverages \acp{LLM} themselves to generate and optimize candidate prompts. This approach allows for the optimization in scenarios where gradient-based techniques are not feasible and has been shown to outperform manual rewriting and few-shot prompting \cite{prasad_2022_grips, zhou_2022_large}. By formulating our objectives in natural language, we leverage the \ac{LLM}'s own capabilities to refine and propose prompts that are likely to yield more accurate and consistent label-rationale pairs.

\begin{figure}[H]
    \centering
    \includegraphics[width=0.62\linewidth]{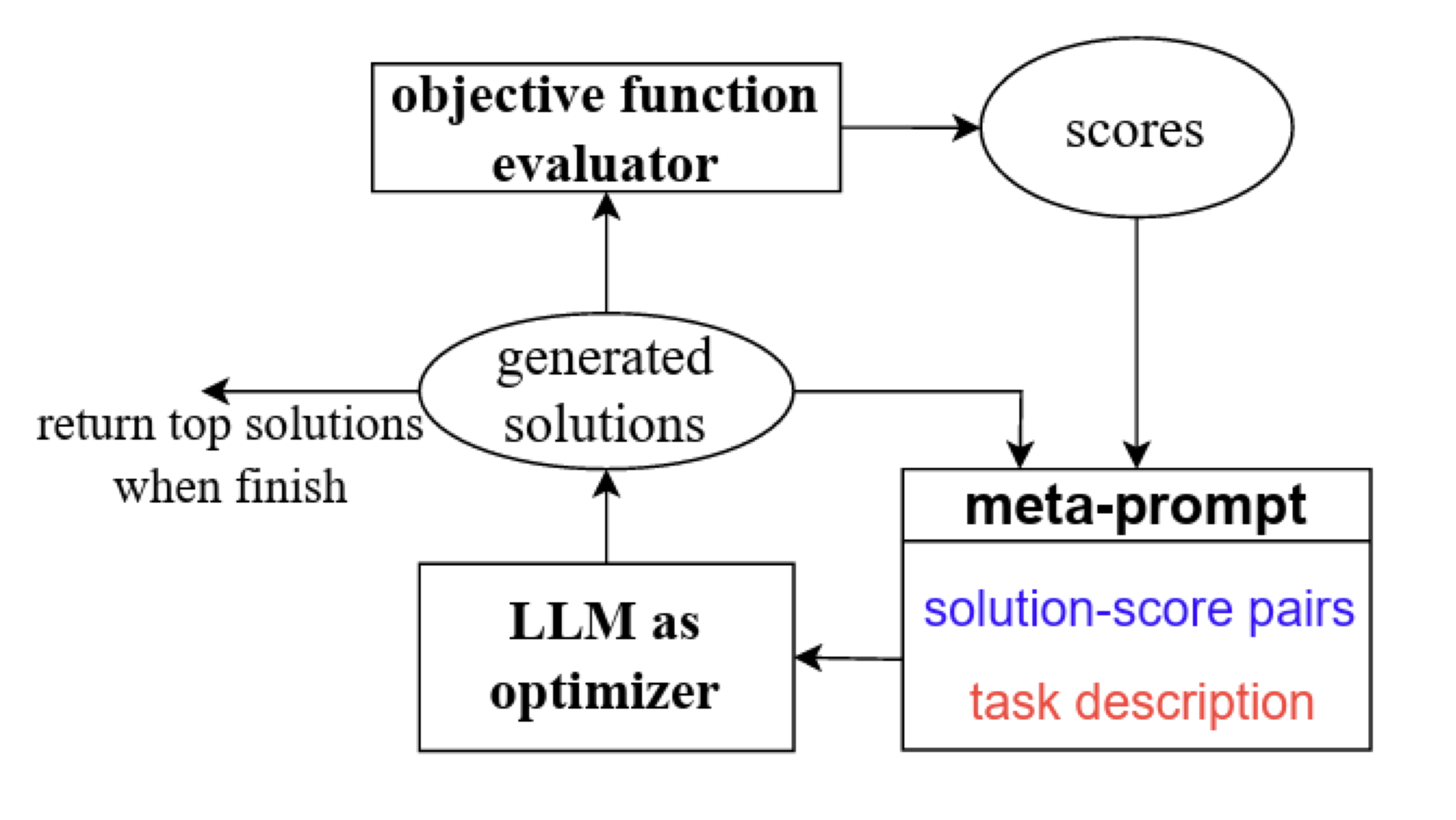}
    \caption{Overview of the OPRO framework by \cite{yang_2023_large}}
\end{figure}

The \ac{OPRO} mechanism functions by iterating over cycles of prompt generation and evaluation. Starting with a meta-prompt that includes a task description and examples of scored solutions from previous iterations, the language model generates new prompt candidates. These candidates are then evaluated based on their ability to elicit accurate labels and coherent rationales from a subset of data. The evaluation process involves parsing the model's responses to separate labels from rationales and comparing these labels against a predefined set of ground truth labels to calculate accuracy. This iterative process encourages the generation of prompts that facilitate easy parsing into labels and rationales while minimizing the need for human intervention.


\section{Experimental Design}

Our experiment is designed to evaluate the efficacy of our zero-shot step-by-step distillation methodology. To this end, we outline the datasets employed to test the generalizability and robustness of our approach, as well as the specifics of the teacher and student models utilized in the distillation process.

\subsection{Models}

We utilize OpenAI's `gpt-3.5-turbo'  as the teacher model, accessed through the OpenAI API on a pay-per-token basis. GPT 3.5 is trained to follow instructions and answer questions in a dialog manner \cite{openai_text}. It usually includes rationales in its responses by default, which we leverage for our distillation process. We conjecture that GPT's optimization for following of zero-shot instructions gives it an edge over other \acp{LLM} in our application.

The student models employed in our study are T5 models of varying sizes (`small' with 77M parameters, `base' with 250M parameters, `large' with 800M parameters, and `XL' with 3B parameters), which have proven their efficacy in transfer learning \cite{raffel_2020_exploring}. Specifically, we use the T5 v1.1 models which were only pretrained on the C4 Common Crawl dataset for general language understanding without additional task-specific finetuning. The use of text-to-text models allows for straightforward processing, as both the labels and rationales can be cast as text sequences, simplifying preprocessing and tokenization.

\subsection{Datasets}

Our main dataset is the first subset of the ANLI database, a collection of \ac{NLI} tasks released by \cite{nie_2019_adversarial}, henceforth referred to as ANLI1. It consists of premise-hypothesis pairs, where the hypothesis either entails, contradicts or is neutral towards the premise and is aimed to benchmark the natural language understanding of \acp{LLM}.

\begin{table}[H]
	\centering
	\caption{Sample of the ANLI1 dataset}
	\scriptsize
    \setlength{\tabcolsep}{6pt}
	{\def\arraystretch{1}
	\begin{tabular}{p{0.55\linewidth}p{0.2\linewidth}p{0.14\linewidth}}
		\toprule
		\textbf{Premise} & \textbf{Hypothesis} & \textbf{Label} \\
		\midrule
		The Parma trolleybus system (Italian: ``Rete filoviaria di Parma'') forms part of the public transport network of the city and ``comune'' of Parma, in the region of Emilia-Romagna, northern Italy. In operation since 1953, the system presently comprises four urban routes. & The trolleybus system has over 2 urban routes. & \textbf{entailment} \\
		\midrule
	    Will Vodery (October 8, 1885 - November 18, 1951) was an African American composer, conductor, orchestrator, and arranger, and one of the few black Americans of his time to make a name for himself as a composer on Broadway, working largely for Florenz Ziegfeld. & Will Vodery wrote a song in 1900. & \textbf{neutral} \\
		\midrule
		The Ten Commandments of Dog Ownership, also known as The Ten Commandments from a Pet's Point of View, or simply Ten Commandments For Dog Owners is a set of pet ownership rules. It was created in 1993 by Stan Rawlinson, and is written from the point of view of a dog. & The Ten Commandments of Dog Ownership were created after 1994. & \textbf{contradiction} \\
		\bottomrule
	\end{tabular}
    }
\end{table}

The second dataset used in our experiment is the \ac{CQA} database by \cite{talmor_2019_commonsenseqa}, a question answering challenge designed to evaluate the ability to reason with commonsense knowledge. It is a collection of multiple-choice questions, that require some form of prior world knowledge to be answered. It therefore does not only test a machine's ability to comprehend the natural language question and answer options, but also its ability to understand concepts that cannot be inferred from the question alone.

\begin{table}[H]
	\centering
	\caption{Sample of the CQA dataset}
	\scriptsize
    \setlength{\tabcolsep}{6pt}
	{\def\arraystretch{1}
	\begin{tabular}{p{0.45\linewidth}p{0.27\linewidth}p{0.17\linewidth}}
		\toprule
		\textbf{Question} & \textbf{Answer options} & \textbf{Label} \\
		\midrule
		``There are 10 apples on an apple tree.  Three fall off.  Now there are X apples.'' What is this an example of? & park,
        coloring book,
        garden center,
        math problem,
        gravity & \textbf{math problem} \\
		\midrule
	    A bald eagle flies over St. Paul, where is it? & texas,
        thermal,
        minnesota,
        canada,
        photograph & \textbf{minnesota} \\
		\bottomrule
	\end{tabular}
    }
\end{table}

Dataset statistics are provided in Table \ref{tab:sizes}. In the case of \ac{CQA}, the validation split is a 10 \% random subsample of the training split, since no dedicated validation split including ground truth labels is available. In both cases, we evaluate the performance of our \acp{LLM} using the accuracy metric.

\begin{table}[H]
	\centering
	\caption{Dataset sizes}
	\scriptsize
    \setlength{\tabcolsep}{6pt}
	{\def\arraystretch{1}
	\begin{tabular}{p{0.1\linewidth}p{0.1\linewidth}p{0.1\linewidth}p{0.1\linewidth}}
		\toprule
		\textbf{Dataset} & \textbf{Train} & \textbf{Test} & \textbf{Eval}\\
		\midrule
		ANLI1 & 16,946 & 1,000 & 1,000 \\
		\midrule
	    CQA & 8,766 & 1,221 & 975 \\
		\bottomrule
	\end{tabular}
    }
	\label{tab:sizes}
\end{table}



\section{Results}

This section presents the findings from our experiments, with each subsection providing different insights into the advantages and implications of our distillation approach. We begin by detailing the results of the prompt optimization process using the \ac{OPRO} mechanism. Following this, we explore the impact of our zero-shot step-by-step distillation approach on the finetuning and distillation accuracy of student models and compare it to few-shot step-by-step distillation. We also analyze the cost efficiency of our approach compared to few-shot methods, highlighting the cost-savings enabled by zero-shot prompting. Finally, we investigate the effects of explanation rates and rationale lengths on student model accuracy.

\subsection{Prompt Optimization}

\begin{figure}[H]
    \centering
    \includegraphics[width=\linewidth]{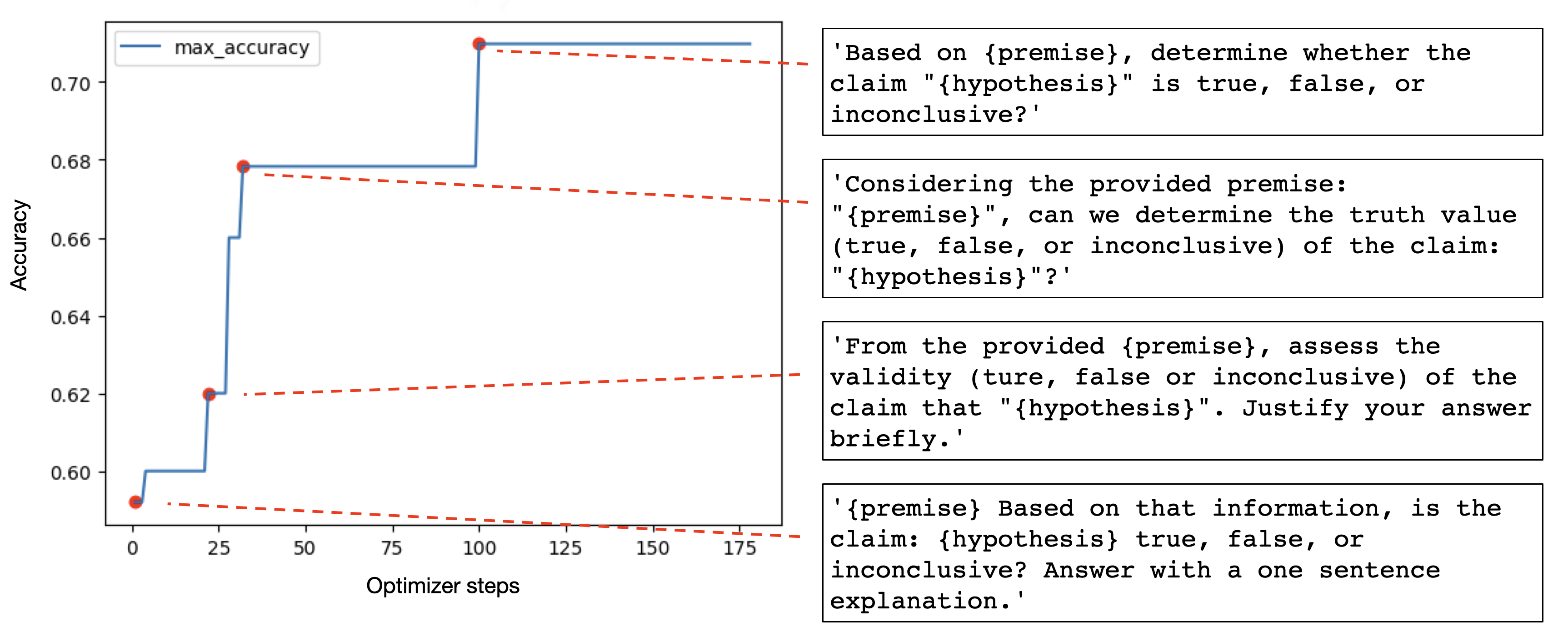}
    \caption{\ac{OPRO} progression on ANLI1}
	\label{fig:opro}
\end{figure}

We ran 22 iterations of \ac{OPRO} in our setup, where each iteration consisted of 8 new candidate prompts being created (querying the meta-prompt 8 times) and evaluated on 25 dataset instances, yielding a total of 176 different prompt templates that have been evaluated. These numbers were chosen based on the results reported in the original paper \cite{yang_2023_large}. The progression of the optimization on the ANLI1 dataset is displayed in Fig. \ref{fig:opro}.

We cannot rule out that more iterations might give an even better prompt template, but the optimization curve is clearly flattening out and the best prompt found already has an accuracy of 71.0 \%, which is better than any benchmark result available for the ANLI1 dataset in current research \cite{laskar_2023_a}. It is also better than the 540B PaLM few-shot accuracy of 70.9 \% achieved by \cite{hsieh_2023_distilling}. The final result of the prompt optimizations and their respective teacher label accuracy and explanation rate are found in Table \ref{tab:opro}.

\begin{table}[H]
	\centering
	\caption{Prompts resulting from OPRO}
	\scriptsize
    \setlength{\tabcolsep}{6pt}
	{\def\arraystretch{1}
	\begin{tabular}{p{0.09\linewidth}p{0.52\linewidth}p{0.1\linewidth}p{0.14\linewidth}}
		\toprule
		\textbf{Dataset} & \textbf{Final prompt} & \textbf{Teacher\newline accuracy} & \textbf{Explanation rate*}\\
		\midrule
		ANLI1 & Based on \{premise\}, determine whether the claim ``\{hypothesis\}'' is true, false, or inconclusive? & 70.98 \% & 87.18 \% \\
		\midrule
	    CQA & Given the following options, what do you think is the correct answer to the question below: \{question\} Options: a) \{choice\_a\} b) \{choice\_b\} c) \{choice\_c\} d) \{choice\_d\} e) \{choice\_e\} Explain your answer with one sentence. & 72.40 \% & 99.40 \% \\
		\bottomrule
	\end{tabular}
    }
	\vspace{0.4em}\\{* Percentage of teacher responses that included a rationale}
	\label{tab:opro}
\end{table}

\subsection{Finetuning and Distillation}\label{sec:r}

Our experiments assess the efficacy of finetuning and distillation processes on student models, using the ANLI1 and CQA benchmark datasets and the 250M T5-Base student model. Our aim is to validate the practical advantages of step-by-step distillation over established methods in general, and the robustness of zero-shot step-by-step learning in particular. We examine setups with access to ground truth labels (finetuning) and without access to ground truth labels (distillation). In either case, we investigate the standard approach (without rationales), the few-shot step-by-step method as per \cite{hsieh_2023_distilling}, and our zero-shot step-by-step approach across different data availability scenarios, i.e., various subsample sizes of the training data.

\begin{figure}[H]
    \centering
    \includegraphics[width=0.95\linewidth]{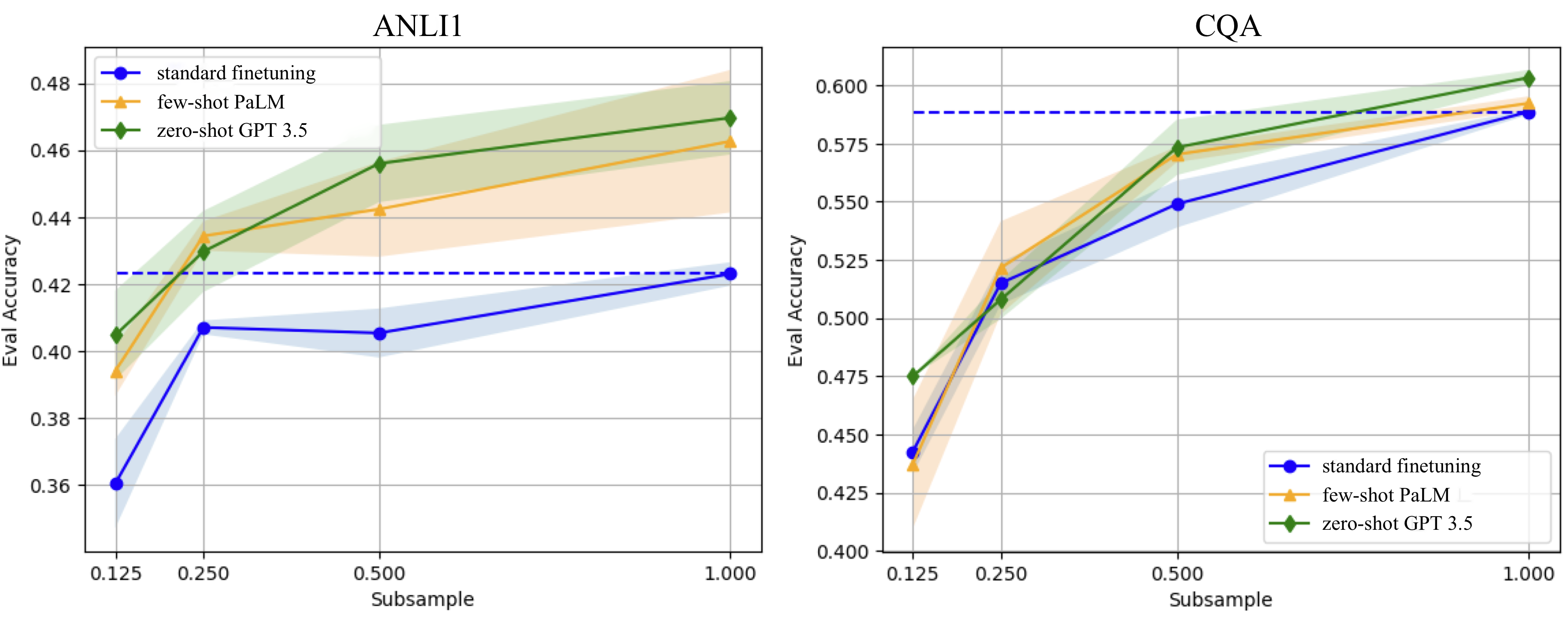}
    \caption{Evaluation accuracy of finetuned student models on varying training set sizes}
	\label{fig:results_finetune}
\end{figure}

Fig. \ref{fig:results_finetune} illustrates that our zero-shot methodology, marked in green, not only aligns with but also occasionally surpasses the performance of the few-shot approach by \cite{hsieh_2023_distilling} in a finetuning scenario. Confidence bars depict the variation in performance across multiple runs with different random weight initializations. The baseline indicated as a dotted line in the plots is the 540B PaLM model finetuned on the entire dataset using only the ground truth labels.

Notably, with less than 25 \% of training data, our method achieves higher accuracy than standard finetuning with the full ANLI1 dataset, indicating significant efficiency in training data utilization. Similarly, for the CQA dataset, our approach demonstrates an advantage over the standard baseline with approximately 75 \% of training data.

An important observation to highlight is that not all generated rationales line up with the correct labeling---only about 71.0 \% for ANLI1 and 72.4 \% for CQA of the explanations are accurate. The remaining rationales, are explanations that lead to a wrong label, since the \ac{LLM} will always produce a rationale
that justifies its prediction, even if it is a wrong prediction. In addition, our optimal prompt has an explanation rate of around 87 \%, meaning that 13 \% of the rationales are mere empty strings.

\begin{figure}[H]
    \centering
    \includegraphics[width=0.5\linewidth]{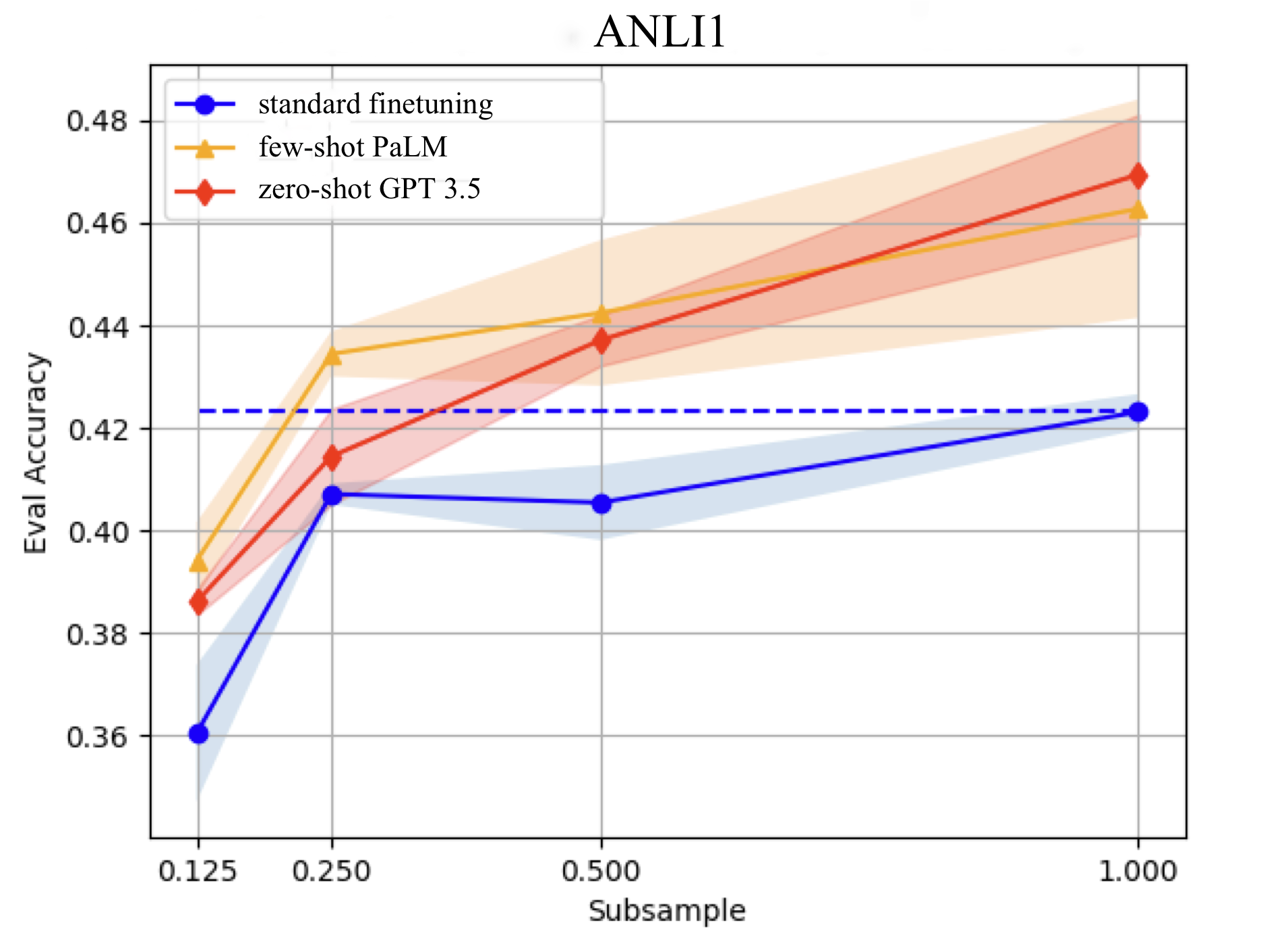}
    \caption{Evaluation accuracy of finetuning with 100 \% correct explanations on ANLI1}
	\label{fig:results_xr}
\end{figure}

To investigate if enhancing the explanation rate and accuracy could elevate the efficacy of step-by-step finetuning even further, we devised teacher prompts that incorporate the correct label directly, nudging the model to only generate rationales justifying why the given label is accurate. The outcomes of employing this revised prompting strategy on the ANLI1 dataset are depicted in Fig. \ref{fig:results_xr}.

Surprisingly, increasing the explanation rate to 100 \% with entirely accurate explanations does not enhance student model performance. Instead, performance slightly declines compared to rationales generated via few-shot prompting, particularly when the student model has access to only a limited portion of the data. It might be that the additional correct explanations display some detrimental intrinsic properties that result from the fact that the teacher model does not understand the reason for the ground truth label on its own.

\begin{figure}[H]
    \centering
    \includegraphics[width=0.97\linewidth]{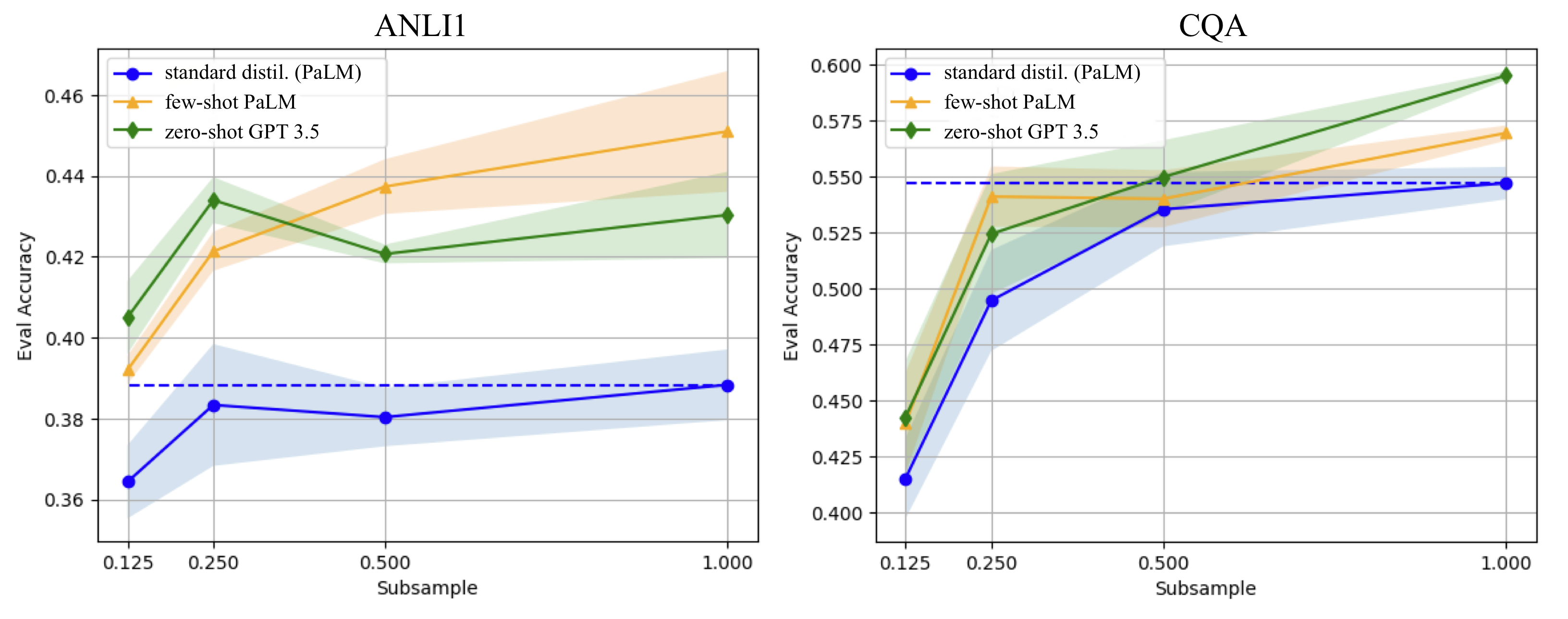}
    \caption{Evaluation accuracy of distilled student models on varying training set sizes}
	\label{fig:results_distil}
\end{figure}

The distillation results, depicted in Fig. \ref{fig:results_distil}, align with the finetuning findings. Step-by-step distillation utilizing zero-shot prompts performs on par with few-shot prompting, affirming the robustness of our approach. Additionally, we observe that step-by-step distillation outperforms the baseline even more than under the assumption of available ground truth labels. Remarkably, with just 12.5 \% of the ANLI1 training data, our approach surpasses the performance of conventional distillation techniques that use the full dataset. This highlights the efficiency of step-by-step learning, especially in contexts where no ground truth labels are available.

\begin{figure}[H]
    \centering
    \includegraphics[width=0.97\linewidth]{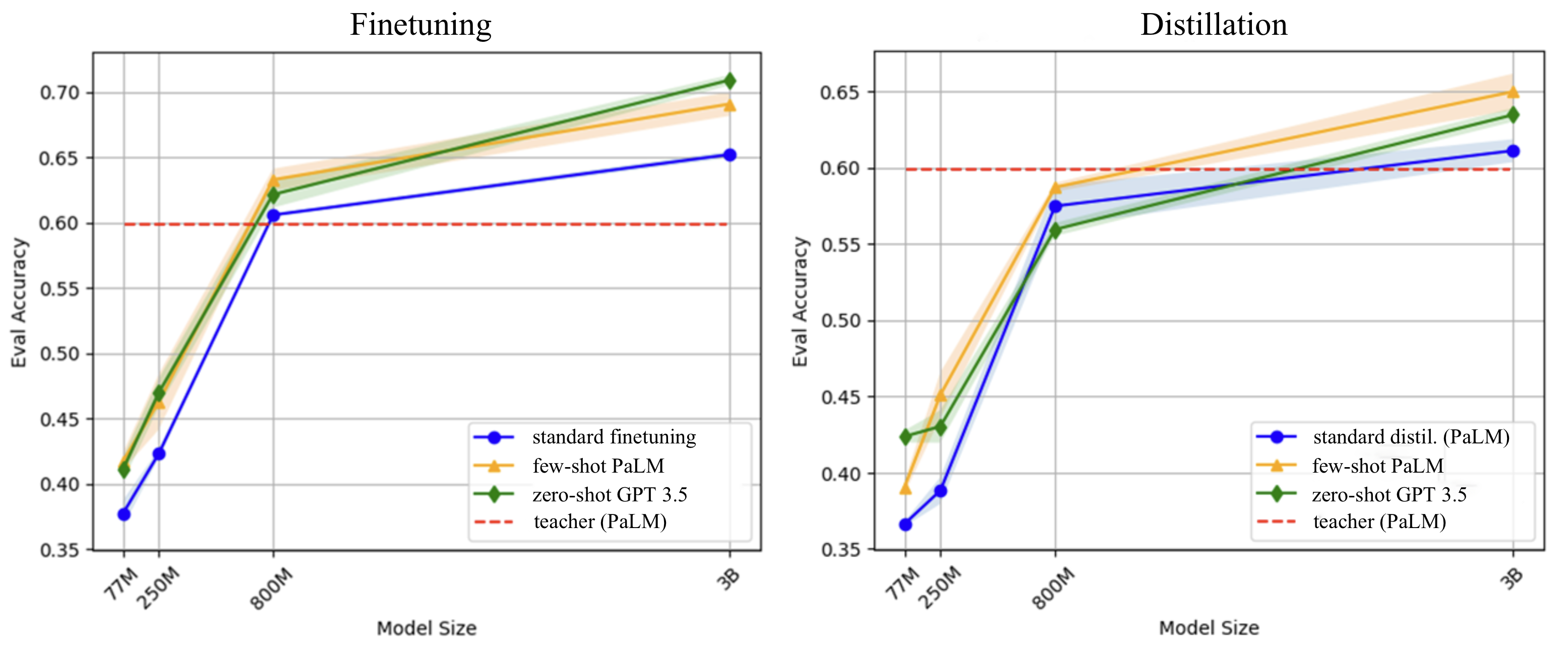}
    \caption{Evaluation accuracy for varying student model sizes}
	\label{fig:results_models}
\end{figure}

Lastly, Fig. \ref{fig:results_models} explores the relationship between student model size and performance. We utilize 100 \% of the available training data and evaluate four T5 model sizes, with 77M, 250M, 800M and 3B parameters. We observe that both few-shot and zero-shot step-by-step approaches improve upon the baseline across all model sizes. The red dotted line showcases the 540B PaLM Teacher \ac{LLM} accuracy. In the case of finetuning, step-by-step learning allows much smaller models, such as the 800M T5-Large, to exceed the teacher \ac{LLM}'s performance, showcasing significant efficiency gains. Notably, the 3B parameter student models sees their performance enhanced by over 9 \%.

These results collectively underscore the effectiveness of zero-shot step-by-step distillation in not only matching but in some instances exceeding the benchmarks set by few-shot methods, while also highlighting the potential for substantial reductions in training data and computational costs.

\subsection{Cost Efficiency}

The results of Section \ref{sec:r} further confirm the main findings of \cite{hsieh_2023_distilling}. Incorporating rationales and labels in step-by-step distillation and finetuning raises effectiveness and efficiency. Student models perform as good or better than the teacher while access to rationales reduces requirements for training data. Furthermore, the results show that our proposition of transitioning from few-shot prompts to more efficient zero-shot prompts does not hurt performance. Specifically, we do not observe a decrease in the quality of relevant properties of the generated explanations or labels.

\begin{figure}[H]
    \centering
    \includegraphics[width=0.77\linewidth]{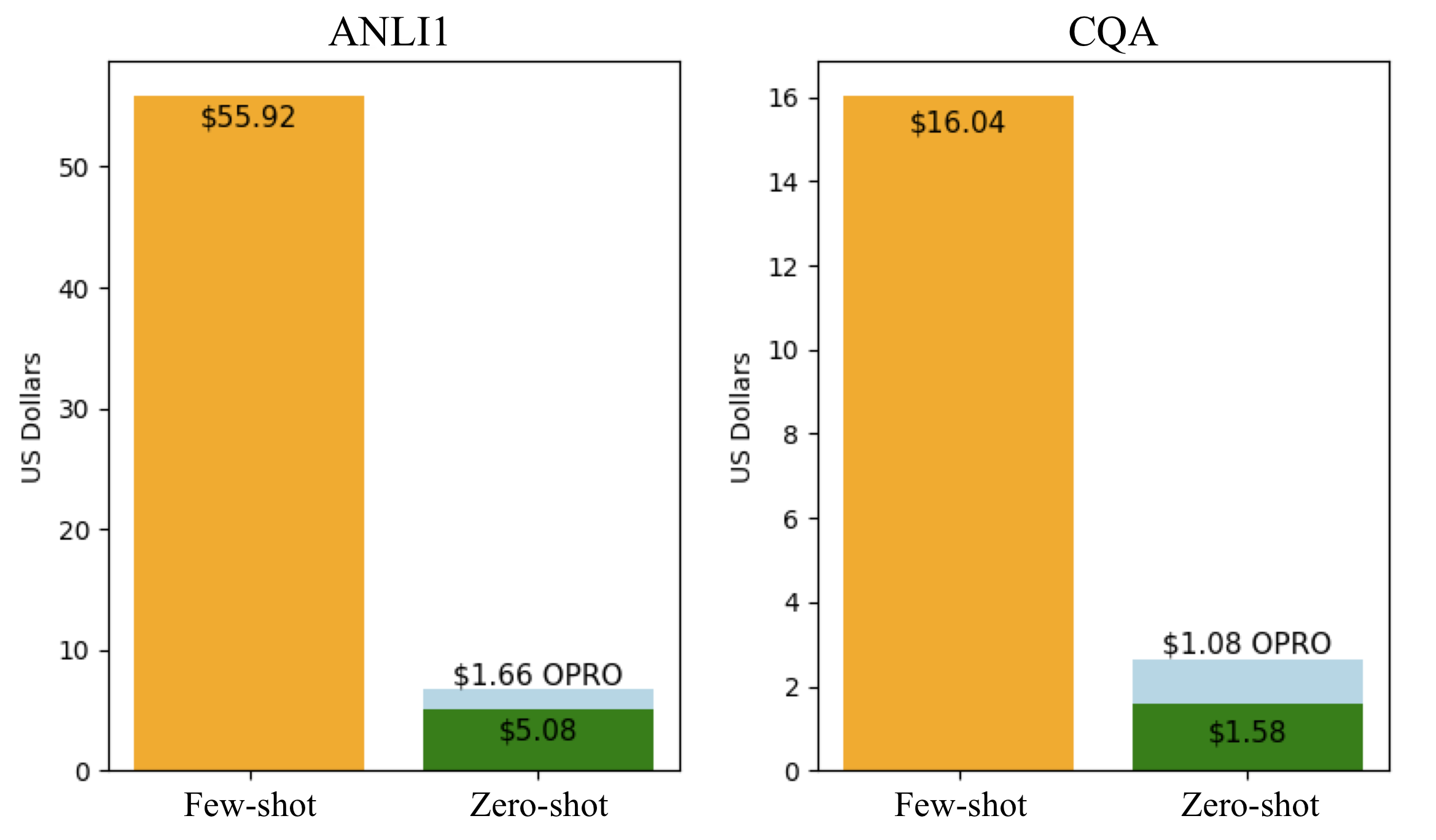}
    \caption{Total costs to annotate the ANLI1 and CQA datasets by prompting approach}
	\label{fig:costs}
\end{figure}

The economic advantages of zero-shot prompting are however substantial as showcased in Fig. \ref{fig:costs}, which compares the annotation costs for the ANLI1 and CQA datasets when using our methodology against the few-shot method employed by \cite{hsieh_2023_distilling}, assuming they were both queried against the same pay-per-token API.

The cost reduction achieved through our zero-shot methodology represents the primary finding of our research and underscores the principal advantage of this approach over conventional few-shot step-by-step distillation. By achieving a tenfold decrease in annotation expenses, our method not only proves its financial efficiency but also its potential to facilitate large-scale \ac{LLM} distillation in a more economically feasible manner.

The costs of the zero-shot approach include the expenditures associated with the \ac{OPRO} prompt optimization process, which also necessitates interactions with the teacher \ac{LLM}. It is important to note that the costs associated with \ac{OPRO} are static, remaining constant regardless of the data volume. This aspect of our approach, especially in the context of growing datasets, positions it as a highly advantageous solution for deploying language models across a wide range of applications where cost and scalability are critical considerations.

\subsection{Effect of Explanation Rate on Student Accuracy}

\begin{figure}[H]
    \centering
    \includegraphics[width=0.97\linewidth]{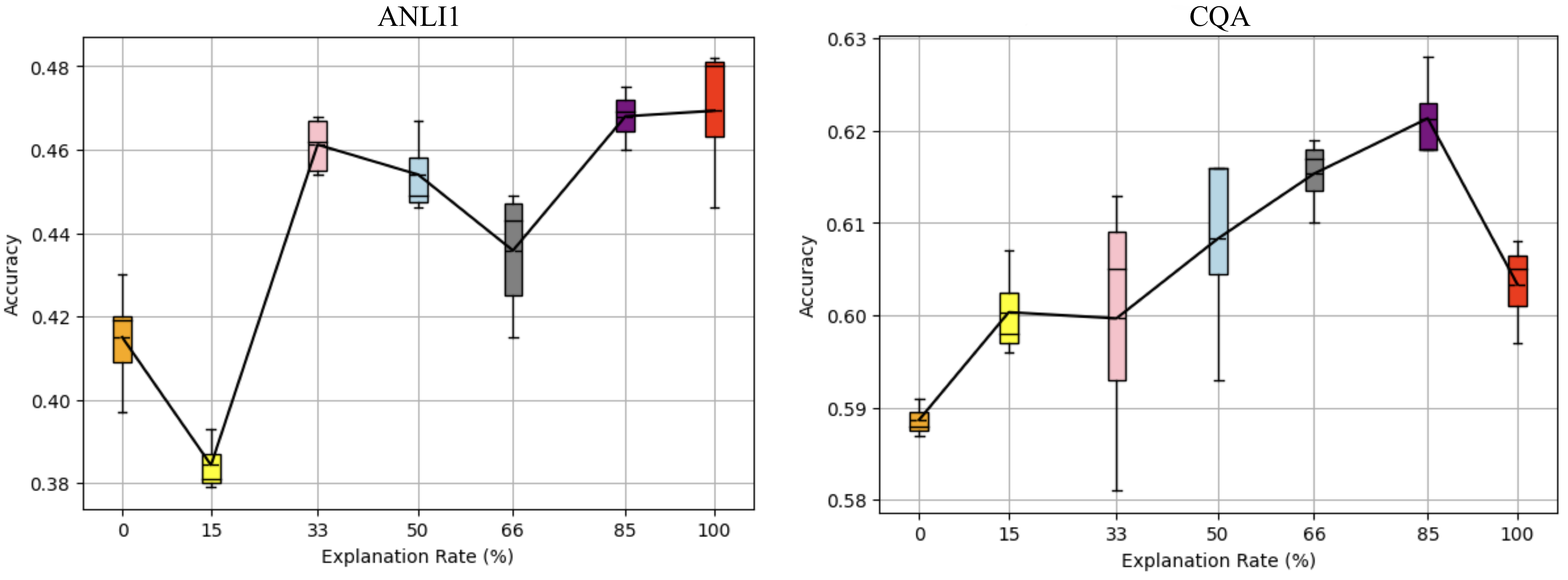}
    \caption{Effect of explanation rate on student model finetuning accuracy}
	\label{fig:xr}
\end{figure}

This section examines how varying explanation rates influence the finetuning accuracy of a 250M T5-Base student model. Explanation rate, defined as the proportion of dataset instances accompanied by a rationale, can vary from 0 \% (standard finetuning without explanations) to 100 \% (each instance includes an explanation). Such analysis helps us understand the balance between rationale inclusion and student model performance. The results are shown in Fig. \ref{fig:xr}, with confidence bars depicting the variation in performance across multiple runs with different random weight initializations.

Our findings reveal a non-linear relationship between explanation rate and student accuracy. Interestingly, a rate of 85 \% equals or surpasses the performance seen at 100 \%, suggesting that beyond a certain threshold, additional rationales do not translate to increased accuracy. This observation underscores the potential for further cost savings without performance loss. Variations across datasets further suggest that optimal explanation rates might depend on specific task characteristics, emphasizing the importance of task-specific optimization in distillation processes.

\subsection{Effect of Explanation Length on Student Accuracy}

\begin{figure}[H]
    \centering
    \includegraphics[width=\linewidth]{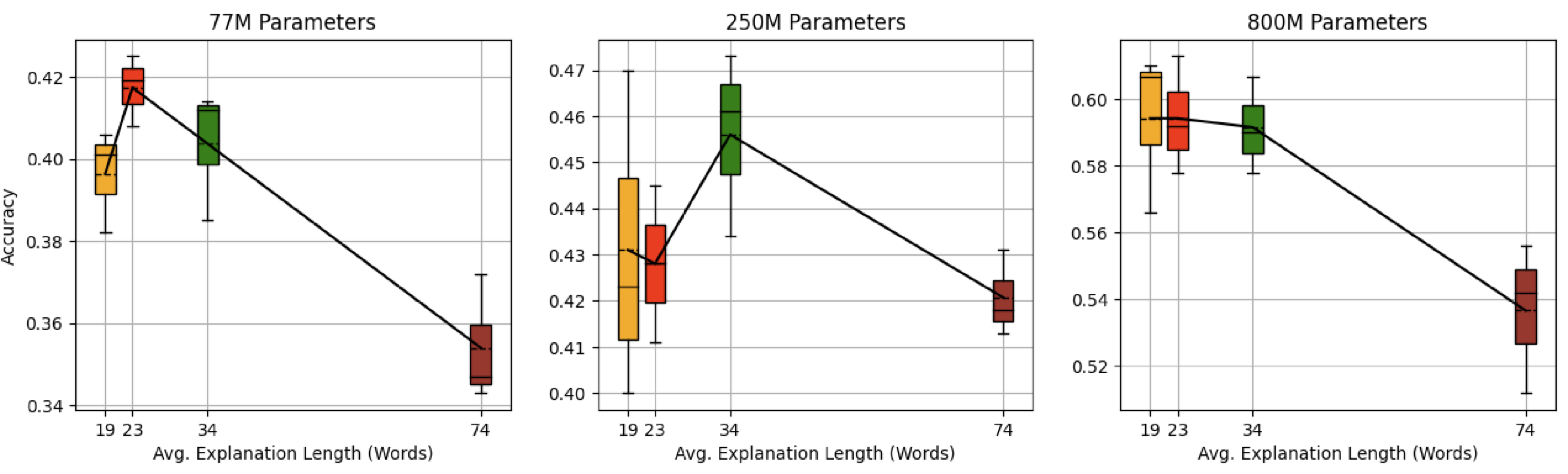}
    \caption{Effect of explanation length and model size on student model accuracy}
	\label{fig:el}
\end{figure}

Following the analysis of the impact of the explanation rate on the accuracy of student models, we delve into the impact of the length of explanations. This part of our study is designed to identify the length of rationales that most effectively enhances student learning outcomes. To achieve variations in the length of explanations provided by the teacher \ac{LLM}, we strategically modify the teacher prompt. Fig. \ref{fig:el} visualizes the accuracy of Small, Base and Large T5 models on the ANLI1 dataset across varying average explanation lengths.

We observe that rationale length, quantified in words, is correlated with the performance of student models. Consistently, across all tested model sizes, we observe a decline in performance with increasing explanation length. The optimal length for the teacher rationale ranges from 20 to 40 words.

The smallest model shows peak accuracy with just 23 words per explanation, whereas the 250M Base model performs best with explanations around 33 words. Interestingly, even the largest model tested does not benefit from longer rationales, echoing the trend observed in smaller models. This pattern underscores a crucial aspect of effective distillation: succinct rationales tend to facilitate more efficient learning than their lengthier counterparts, likely due to the focused and concise transfer of knowledge they enable.


\section{Conclusion}

Our study introduces a novel approach to distilling \acp{LLM}, demonstrating the practicality and efficiency of zero-shot step-by-step distillation over traditional few-shot methods. By sidestepping the need for handcrafted few-shot examples, we cut down annotation costs by approximately 85 \%. This cost reduction, achieved without sacrifices in performance, underscores the viability of zero-shot distillation for industrial applications, offering a scalable, cost-effective solution for deploying powerful, task-specific models with minimal human intervention.

Key findings reveal that annotating a dataset entirely with explanations is unnecessary for achieving near-optimal performance gains. Specifically, providing explanations for about a third of the dataset can secure over 90 \% of the potential performance improvement, suggesting a cost-effective task-dependent optimization strategy. The negligible impact of incorrect explanations on model performance underlines the robustness of the zero-shot distillation process, attributed to the training's design, which separates the predictions of labels and rationales. Emphasis on concise explanations can further enhance model performance and reduce costs.

This research confirms the potential for significant cost savings and operational efficiencies in the training of LLMs, thereby encouraging the adoption of zero-shot step-by-step distillation as a standard practice in the development of customized artificial intelligence models. By highlighting the effectiveness of selective rationale annotation and the benefits of concise explanations, it also sets the stage for further explorations into the optimization of distillation processes.


\begin{credits}

\subsubsection{\discintname}
The authors have no competing interests to declare that are
relevant to the content of this article.

\end{credits}


\newpage

\bibliographystyle{splncs04}
\bibliography{bibliography.bib}


\end{document}